\documentclass{article}
\usepackage{xcolor}
\newcommand{\jmf}[1]{}\newcommand{\jmfnew}[1]{}\newcommand{\zekun}[1]{}  %

\usepackage{microtype}
\usepackage{graphicx}
\usepackage{booktabs}
\usepackage{placeins}
\usepackage{hyperref}

\makeatletter\@namedef{ver@algorithmic.sty}{}\makeatother

\usepackage[accepted]{mlsys2025}
\usepackage{algpseudocode}

\usepackage{amsmath,amsfonts,bm}

\def\eqref#1{equation~\ref{#1}}

\def\1{\bm{1}}

\DeclareMathAlphabet{\mathsfit}{\encodingdefault}{\sfdefault}{m}{sl}
\SetMathAlphabet{\mathsfit}{bold}{\encodingdefault}{\sfdefault}{bx}{n}

\usepackage{booktabs}       %
\usepackage{amsfonts}       %
\usepackage{nicefrac}       %
\usepackage{xcolor}         %

\usepackage{algpseudocode}

\usepackage{inconsolata}
\usepackage{amsmath}

\usepackage{algpseudocode}
\usepackage{comment}
\usepackage{multirow}
\usepackage{amsfonts}       %
\usepackage{nicefrac}       %

\usepackage{amsthm}
\newtheorem{theorem}{Theorem}

\newtheorem{proposition}[theorem]{Proposition}
\usepackage{subcaption}
\DeclareMathOperator{\softmax}{softmax}
\newcommand{\R}{\mathbb{R}}

\theoremstyle{plain}
\theoremstyle{definition}

\theoremstyle{remark}

\mlsystitlerunning{Read Once, Answer Many: Parallel Prompting for Shared-Context LLM Inference}

\begin{document}

\twocolumn[
\mlsystitle{Read Once, Answer Many:\\Parallel Prompting for Shared-Context LLM Inference}

\mlsyssetsymbol{equal}{*}

\begin{mlsysauthorlist}
\mlsysauthor{Zekun Zhao}{ucsc}
\mlsysauthor{Jeffrey Flanigan}{ucsc}
\end{mlsysauthorlist}
\mlsysaffiliation{ucsc}{University of California, Santa Cruz}

\mlsyskeywords{LLM inference, shared prefix, KV cache, attention kernels, SGLang}

\vskip 0.3in

\begin{abstract}
LLM workloads sometimes ask many independent questions about the same long document, as in document QA, data labeling and LLM-based evaluation.
Prefix caching removes the redundant prefill shared by these queries, but each decode step still reads the document's key--value (KV) cache once per question, so decoding is bound by memory bandwidth and scales with the number of questions.
We present \emph{Parallel Prompting}, which decodes $P$ questions as $P$ independent streams of a single request all at once, fixing this issue. In our method, every step reads the shared document once for all $P$ streams, and a stream-id attention mask, evaluated inside the attention kernel, keeps each answer conditioned only on the document and its own question.
We implement Parallel Prompting in SGLang by adding about 80 lines to three existing files and a small engine package; the mask needs no dense tensor, and every decode step has a fixed shape that is captured as a CUDA graph.
With CodeLlama-7B, 8k--16k-token documents and 64--256 questions per document, Parallel Prompting is \textbf{3.7$\times$ faster} than stock SGLang with its FlashInfer backend on an A100 and \textbf{5.9--6.0$\times$ faster} on an RTX A6000, on two SGLang releases, including the most recent release. Timed on identical inputs, it is faster than stock SGLang in all 144 runs (2 GPUs $\times$ 2 releases $\times$ 6 workloads $\times$ 6 input sets).
Code: \url{https://github.com/jlab-nlp/parallel-prompting}. Project page: \url{https://jlab-nlp.github.io/parallel-prompting/}.
\end{abstract}
]

{\let\thefootnote\relax\footnotetext{\hspace*{-\footnotesep}\textsuperscript{1}University of California, Santa Cruz. Preprint.}}

\section{Introduction}
\label{sec:intro}

Large language models (LLMs) power chat assistants, coding agents, analysis engines and domain-specific agents, and the cost of inference has become a major bottleneck for deploying them~\citep{kwon2023efficient}.
Many of these requests share a large part of their input, a \emph{shared prefix}: long system prompts~\citep{zhu2024relayattention}, safety policies~\citep{inan2023llamaguard}, retrieved documents~\citep{lewis2020retrieval}, few-shot exemplars or domain knowledge~\citep{wu2023bloomberggpt}.
In document question answering, large-scale labeling and LLM-based evaluation, tens to hundreds of questions are asked about the same long document~\citep{pang2022quality,wang2021want,gilardi2023chatgpt,zheng2023judging,liu2023geval}.

Existing serving engines already exploit this structure.
PagedAttention~\citep{kwon2023efficient} and SGLang's RadixAttention~\citep{zheng2024sglang} store the document's KV cache once and share it across requests, so the document is prefilled only once.
Decoding, however, is still the bottleneck: each question is its own request, and each decode step of each request reads the \emph{whole} document's KV cache from GPU memory again.
With $Q$ questions about a document of $L$ tokens, every decode step therefore moves $Q \cdot L$ tokens' worth of KV, and the step is limited by memory bandwidth rather than compute.
An alternative, \emph{batch prompting}~\citep{cheng2023batch,lin2024batchprompt}, puts all questions into one prompt and decodes all answers in one sequence.
It reads the document once, but the answers attend to each other, which causes cross-question interference and can degrade quality.

We propose \textbf{Parallel Prompting}, which keeps the efficiency of batch prompting and the independence of per-question requests.
It places $P$ questions in one request and decodes them as $P$ \emph{streams}: each step appends one token per stream, and a stream-id mask lets a token attend only to the document and to earlier tokens of its own stream.
Each stream's positions start right after the document, as if its question were alone.
The document's KV is then read roughly once per request per step instead of once per question, which turns $P$ matrix--vector attention products into one matrix--matrix product.

We build Parallel Prompting into SGLang~\citep{zheng2024sglang}.
The stream-id mask is evaluated inside SGLang's existing Triton~\citep{tillet2019triton} extend-attention kernel from two fixed-size index tensors, so no dense mask is built and every decode step is captured as a CUDA graph.
The change adds about 80 lines to three SGLang files plus a small engine package, and we ported it to two releases.
Against stock SGLang with prefix caching and CUDA graphs, Parallel Prompting is 3.7$\times$ faster than the FlashInfer~\citep{ye2025flashinfer} backend on an A100 and 5.9--6.0$\times$ faster on an RTX A6000, and 5.1--7.5$\times$ faster than the Triton backend (\autoref{fig:sglang-portability}).
These figures are for CodeLlama-7B, which uses multi-head attention; on grouped-query-attention models the gain is 1.3--1.9$\times$ (\autoref{sec:eval-models}).

Our contributions are:
\begin{itemize}
    \item We propose \emph{Parallel Prompting}, which decodes many questions about one document as independent streams of a single request (\autoref{sec:background} and \autoref{sec:design}).
    \item We implement it in a production engine, SGLang, with an in-kernel stream-id mask that stays compatible with CUDA graphs, and port it across two releases (\autoref{sec:impl}).
    \item We evaluate it with paired, seeded measurements on two GPUs, two SGLang versions and six models. On the main grid it is faster in all 144 runs, each timed on the same inputs as stock SGLang. We also show how the gain depends on document length and question count, how it shrinks with long answers, and how to choose $P$(\autoref{sec:eval}).
\end{itemize}

\begin{figure*}[t]
\centering
\includegraphics[width=\textwidth]{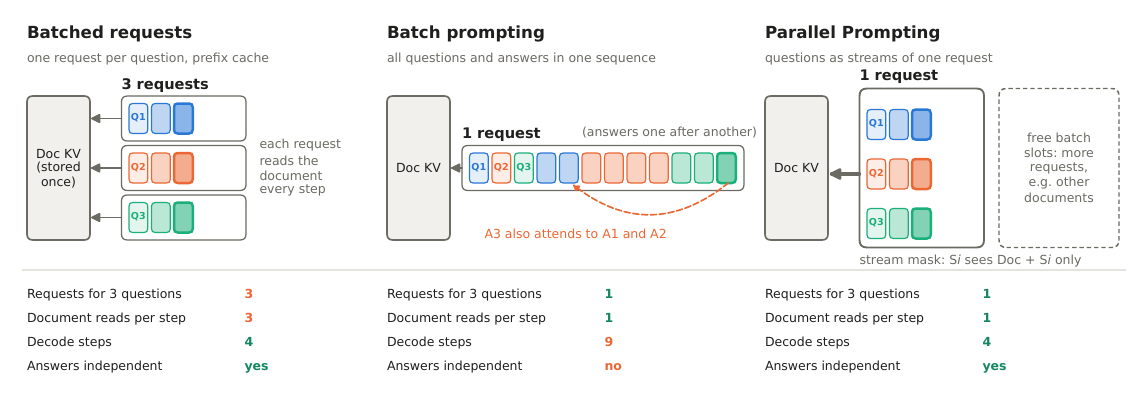}%
\caption{Three ways to answer several questions about one document, shown at one decode step (dark: the token generated in this step). A serving engine such as SGLang or vLLM runs a \emph{batch} of \emph{requests}, and every decode step adds one token to each request. \textbf{Batched requests} with prefix caching store the document's KV once but use one request per question, so every step reads the document once per question. \textbf{Batch prompting}~\citep{cheng2023batch,lin2024batchprompt} puts all questions and answers in one sequence: it reads the document once, but the answers are generated one after another, and later answers attend to earlier ones. \textbf{Parallel Prompting} decodes the questions as \emph{streams} of one request: every step adds one token per stream and reads the document once for all of them, and a stream mask (\autoref{fig:stream-mask}) keeps the answers independent. The batch still has room for other requests, for example for other documents (\autoref{fig:sglang-groups}). }
\label{fig:3step}
\end{figure*}

\section{Background and Motivation}
\label{sec:background}

\paragraph{Attention in prefill and decode.}
Given queries $\mathbf{Q} \in \R^{N_q \times d}$, keys $\mathbf{K} \in \R^{N_{kv} \times d}$ and values $\mathbf{V} \in \R^{N_{kv} \times d}$, attention computes
\begin{align}
    \mathbf{O} = \softmax{\left( \frac{\mathbf{Q}\mathbf{K}^T} {\sqrt{d}} + \mathbf{M} \right)}\mathbf{V},
    \label{eq:attention}
\end{align}
with a causal mask $\mathbf{M}$.
During prefill, the whole prompt is processed in one forward pass ($N_q = N_{kv} \gg 1$), and attention is a compute-friendly matrix--matrix product.
During decode, each step processes one new token per sequence against a KV cache of all earlier tokens ($N_q = 1$, $N_{kv} \gg 1$).
Attention becomes a matrix--vector product whose cost is dominated by reading $\mathbf{K}$ and $\mathbf{V}$ from memory.
Batching many sequences helps the MLP layers, which share weights, but not attention, because each sequence has its own KV cache.

\paragraph{Shared prefixes under prefix caching.}
Consider a document $D$ of $L$ tokens and questions $q_1, \ldots, q_Q$, where each answer should be $a_i = \pi_{\text{LLM}}(D, q_i)$.
A serving engine with prefix caching prefills $D$ once and points every request's KV table at the same cached pages.
This saves compute and memory, but each decode step of each request still attends over all $L$ document tokens.
A step over $Q$ requests therefore reads the document's KV $Q$ times.
For long documents this term dominates the memory traffic of a decode step, and it grows linearly in both $L$ and $Q$.
Shared-prefix attention kernels~\citep{juravsky2024hydragen,zhu2024relayattention,ye2024chunkattention,athiwaratkun2024bifurcated,ye2025flashinfer,pan2025fasttree} attack this term with a dedicated kernel that computes the prefix part of attention for many requests at once and merges it with each request's suffix attention.

\paragraph{Batch prompting.}
Batch prompting~\citep{cheng2023batch,lin2024batchprompt} concatenates all questions after $D$ and lets the model write all answers in one sequence.
It reads $D$ once per step, but the hidden state for answer $i$ contains every earlier question and answer.
Answers are no longer independent, and quality can degrade; our prototype shows this on SQuAD and DROP (\autoref{tab:empirical-table-horizontal} in \autoref{app:hf}).
 \autoref{fig:3step} contrasts batch prompting with batched requests and with Parallel Prompting.

\section{Parallel Prompting}
\label{sec:design}

Parallel Prompting generates all answers in parallel while keeping each answer conditioned only on $(D, q_i)$.
A request holds the document followed by $P$ questions, and is decoded as $P$ streams:
\begin{center}
\small
\texttt{[ D ]
[ q$_1$ ]\ldots[ q$_P$ ][ a$_1^1$ \ldots\ a$_P^1$ ][ a$_1^2$ \ldots\ a$_P^2$ ]\ldots}
\end{center}
Here $a_s^t$ is the $t$-th token of stream $s$.

\paragraph{Stream ids and the mask.}
Every KV slot carries a stream id $\mathrm{sid}$: $-1$ for document tokens and $s$ for tokens of question $s$ and its answer.
A token of stream $s$ may attend to a key $k$ only if
\begin{equation}
\mathrm{causal}(q,k) \;\wedge\; \bigl(\mathrm{sid}_k = -1 \;\vee\; \mathrm{sid}_k = \mathrm{sid}_q\bigr).
\label{eq:streammask}
\end{equation}
Every stream therefore sees exactly the tokens it would see if its question were decoded alone.

\paragraph{Positions.}
Each stream's positions start at $L$, right after the document, so rotary embeddings match those of a single-question request.

\paragraph{Prefill.}
If $D$ is already cached, only the $P$ questions are prefilled, as one packed sequence under \autoref{eq:streammask}, in the manner of prepacking~\citep{zhao2024prepacking}.
Otherwise $D$ is prefilled once and shared by every request that uses it.

\paragraph{Parallel decode.}
Each decode step appends one token to every stream: $P$ query tokens per request, all attending to the same document KV.
The document's keys and values are loaded once per query tile, instead of once per question.
The $P$ matrix--vector products of standard decoding thus become one matrix--matrix product with $P$ rows.
The output logits are sampled independently, and finished streams are masked out until the request ends.

\paragraph{Batching and choosing $P$.}
Questions about one document are split into $\lceil Q/P \rceil$ requests, and requests over different documents are batched together as usual.
$P$ trades attention efficiency against the extra keys each stream must skip: the KV row of a request also holds the other streams' tokens, which the kernel scans and masks out.
Our analysis in \autoref{app:theory} suggests that throughput is maximized at an interior point.
Writing $B$ for the batch size, the number of requests per step, throughput is
\begin{equation}
\mathrm{Thr}(P,B) = \frac{P\,B}{T_{\mathrm{attn}}(P) + T_{\mathrm{mlp}}(P B)},
\label{eq:thr}
\end{equation}
where $T_{\mathrm{attn}}$ falls with $P$ until the extra masked keys dominate, and $T_{\mathrm{mlp}}$ grows with the number of tokens, but sublinearly until the GPU is saturated.
In particular, the largest $P$ does not maximize \autoref{eq:thr}.
In \autoref{sec:eval-p}, we show that the optimum is broad and similar across GPUs and engine versions.
\autoref{alg:prompting} summarizes the generation loop.

\begin{algorithm}[!t]
\footnotesize
\caption{Parallel Prompting generation loop (prototype formulation with explicit masks; the SGLang engine evaluates the mask inside the kernel, Section~\ref{sec:impl}).}
\begin{algorithmic}[1]

\Require Shared prefix \texttt{Doc}, unique suffix set \texttt{Q\_all}, questions per batch \texttt{N}, streams per request \texttt{P}, language model \texttt{$\pi_{\text{LLM}}$}
\Ensure List of generated answers

\State Optional: \texttt{cache} $\gets$ \textsc{Precompute}(\texttt{$\pi_{\text{LLM}}$}, \texttt{Doc})
    \Comment{Prefill KV-cache for shared prefix}

\State \texttt{i} $\gets 0$
\State \texttt{N\_p} $\gets$ \texttt{N} / \texttt{P}  \Comment{Requests per batch ($B$)}

\While{\texttt{i} $<$ $|\texttt{Q\_all}|$}
    \State \texttt{Q\_n} $\gets$ \texttt{Q\_all}[\texttt{i} : \texttt{i + N}]
    \State \texttt{Q\_np} $\gets$ \textsc{ParallelizeInterleave}(\texttt{Q\_n}, \texttt{P})
    \State \texttt{prompts} $\gets$ \textsc{PrepareInput}(\texttt{Doc}, \texttt{Q\_np}, \texttt{N\_p})
    \State \texttt{masks} $\gets$ \textsc{PrepareMask}(\texttt{prompts})

    \State \texttt{answers}, \texttt{output\_mask} $\gets$
        \textsc{ParallelGenerate}(\texttt{$\pi_{\text{LLM}}$}, \texttt{prompts}, \texttt{masks}, \texttt{P}, \texttt{cache})

    \For{\texttt{n} = 1 \textbf{to} \texttt{N\_p}}
        \For{\texttt{p} = 1 \textbf{to} \texttt{P}}
            \State \texttt{final\_answer.append(}\textsc{Decode}(\texttt{answers[n, p]}, \texttt{output\_mask[n, p]})\texttt{)}
        \EndFor
    \EndFor

    \State \texttt{i} $\gets$ \texttt{i + N}
\EndWhile

\State \Return \texttt{final\_answer}

\\
\Function{ParallelGenerate}{\texttt{$\pi_{\text{LLM}}$}, \texttt{prompts}, \texttt{masks}, \texttt{P}, \texttt{cache}}
    \State \texttt{finished} $\gets$ \texttt{False}
    \State \texttt{input\_ids} $\gets$ \textsc{Tokenize}(\texttt{prompts})

    \While{\textbf{not} \texttt{finished}}
        \State \texttt{outputs} $\gets$
            \texttt{$\pi_{\text{LLM}}$}.\textsc{Forward}(\texttt{input\_ids}, \texttt{masks}, \texttt{cache})

        \State \texttt{logits} $\gets$ \texttt{outputs[:,\, -P:]} \Comment{Outputs \texttt{P} logits on sequence dimension}
        \State \texttt{next\_tokens} $\gets$ \textsc{Sample}(\texttt{logits})
        \State \texttt{input\_ids} $\gets$ \textsc{Concat}(\texttt{input\_ids}, \texttt{next\_tokens})

        \If{\textsc{StoppingCriteria}(\texttt{input\_ids})}
            \State \texttt{finished} $\gets$ \texttt{True}
        \Else
            \State \texttt{masks} $\gets$ \textsc{UpdateParallelMask}(\texttt{input\_ids}, \texttt{P})
        \EndIf
    \EndWhile

    \State \Return \texttt{input\_ids}, \texttt{masks}
\EndFunction

\\
\Function{Precompute}{\texttt{$\pi_{\text{LLM}}$}, \texttt{Doc}}
    \State \texttt{kv\_cache} $\gets$ \texttt{$\pi_{\text{LLM}}$}.\textsc{Forward}(\texttt{Doc})
    \State \Return \texttt{kv\_cache}
\EndFunction

\end{algorithmic}
\label{alg:prompting}
\end{algorithm}

\section{Implementation in SGLang}
\label{sec:impl}

We implemented Parallel Prompting in SGLang v0.5.22 and v0.4.2.post3.
\autoref{tab:diff} lists our changes.
Every edit is inactive unless a Parallel Prompting batch turns it on, so stock SGLang takes its original code path.

\begin{table}[t]
\centering
\small
\caption{Modifications based on SGLang v0.5.22. The three edited files gain 79 lines.}
\label{tab:diff}
\footnotesize
\begin{tabular}{@{}lp{3.3cm}r@{}}
\toprule
File & Change & Lines \\
\midrule
\texttt{extend\_attention.py} & stream-id mask in both kernel stages; small tile for decode steps & +50 \\
\texttt{triton\_backend.py} & pass \texttt{q\_sid}, \texttt{slot\_sid} to the kernel & +11 \\
\texttt{logits\_processor.py} & one logit row per stream (\texttt{logit\_index}) & +18 \\
\texttt{parallel\_prompt/} & new: layout, reference mask, engine & new \\
\bottomrule
\end{tabular}
\end{table}

\begin{figure*}[t]
\centering
\includegraphics[width=0.85\textwidth]{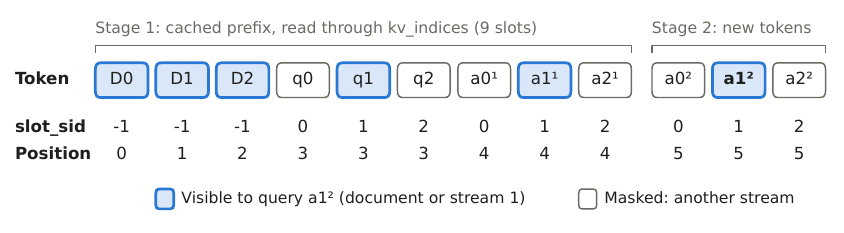}
\caption{The in-kernel mask in one decode step: a request with a 3-token document, $P = 3$ streams and 1-token questions, at decode step 2 ($a_i^{k}$ is stream $i$'s $k$-th answer token). Each KV slot carries its stream id $\mathrm{slot\_sid}$ ($-1$ for the document), and the streams share positions. The new token $a_1^2$ of stream 1 attends only to the document and to stream 1's slots. Stage 1 of the kernel reads the cached slots through \texttt{kv\_indices}; stage 2 reads the step's new tokens.}
\label{fig:stream-mask}
\end{figure*}

\paragraph{In-kernel mask.}
SGLang's Triton \emph{extend} kernel attends a batch of new tokens to cached tokens (stage 1) and to themselves (stage 2).
We add two integer tensors: $\mathrm{q\_sid}$, the stream id of each new token, and $\mathrm{slot\_sid}$, the stream id of each KV slot.
In both stages, the kernel loads the two tensors and combines the condition of \autoref{eq:streammask} with its causal mask (\autoref{fig:stream-mask}).
A dense $\text{tokens} \times \text{KV}$ mask would grow with the KV row and change size every step.
The two index tensors have a fixed size per step, so a step can be captured as a CUDA graph.
For decode steps, which have only $P$ new tokens per request, we use a smaller query tile.

\paragraph{Engine.}
Parallel Prompting runs as an engine on SGLang's \texttt{ModelRunner}.
It prefills each document once, then gives each request a row in the request-to-token table that begins with the document's cached slots, so the document is shared without copying.
Questions are split into requests of $P$ streams and run in batches sized to the free KV pool.
Every step, prefill or decode, is an ordinary EXTEND batch.
One CUDA graph is captured for each pair $(B, P)$, where $B$ is the number of requests in a batch and $P$ the number of streams per request. $B$ is rounded to a power of two, so batches of similar size reuse one graph.
The logits processor returns the $P$ last-token rows of each request, one per stream.
In terms of \autoref{alg:prompting}, \textsc{PrepareMask} becomes the two index tensors, and each \textsc{Forward} call becomes one EXTEND step over all requests of a batch.

\paragraph{Continuous batching.}
By default, the engine runs each batch to completion, so a batch waits for its slowest stream.
In the continuous-batching variant, the engine checks for early finished requests and never lets a request pass its answer budget.
A request whose streams have all stopped leaves the batch, its slots return to the KV pool, and waiting requests are prefilled into the free places.
Each request keeps its own step counter, so requests in one batch can be at different answer positions. The batch is padded to a power of two with a dummy request, so the CUDA graph for each $(B, P)$ is reused.

\paragraph{Correctness checks.}
A CPU test compares the layout and positions with a reference implementation of \autoref{eq:streammask}.
A kernel test compares the Triton output with PyTorch scaled-dot-product attention under the dense reference mask.
On three layouts (documents of 37, 300 and 1{,}000 tokens with 1 to 64 streams), covering prefill and four decode steps, the largest absolute error in bf16 is 0.009, below the test's tolerance of 0.03.
An end-to-end check compares Parallel Prompting's answer quality with stock SGLang's (\autoref{sec:eval-quality}).

\section{Evaluation}
\label{sec:eval}

We compare Parallel Prompting with stock SGLang and answer four questions:
\begin{enumerate}
    \item How much faster is Parallel Prompting than a tuned production engine (\autoref{sec:eval-main})?
    \item Which workloads benefit, and which do not (\autoref{sec:eval-scaling})?
    \item How should the number of streams per request $P$ be chosen (\autoref{sec:eval-p})?
    \item How fast and how accurate is it on real long-document benchmarks, with prefix caching and continuous batching (\autoref{sec:eval-real})?
\end{enumerate}
We then repeat the main experiments on other models (\autoref{sec:eval-models}) and summarize the results of our HuggingFace prototype (\autoref{sec:eval-proto}).

\subsection{Setup}
\label{sec:eval-setup}

\paragraph{Systems.}
We compare three configurations of the same SGLang build on the same GPU:
\begin{itemize}
    \item \textbf{Stock FlashInfer:} stock SGLang with prefix caching, the FlashInfer attention backend~\citep{ye2025flashinfer} and CUDA graphs. It is SGLang's default backend and the faster stock configuration in most of our settings (\autoref{sec:eval-main}).
    \item \textbf{Stock Triton:} the same engine with the Triton backend. It shares the attention backend with Parallel Prompting, but stock decoding uses the backend's decode kernel, while Parallel Prompting runs every step through the extend kernel.
    \item \textbf{Parallel Prompting (PP):} our engine on the Triton backend with CUDA graphs.
\end{itemize}
For the stock configurations, the document is cached before timing, and CUDA graphs are captured for the full decode batch.
We raise SGLang's CUDA-graph batch-size limit to the full decode batch, because the default limit can be smaller, which would understate stock performance.
On the real-data benchmarks (\autoref{sec:eval-real}), every configuration uses SGLang's radix prefix cache. \textbf{PP} also inserts each document into the radix cache, so a prefix shared by several documents, such as the instruction, is prefilled once. A separate study measures \textbf{PP + continuous batching (PP+CB)}, which adds the scheduling described in \autoref{sec:impl}: requests leave the batch as they finish, and waiting requests take their places.

\paragraph{Hardware and software.}
We use an NVIDIA A100-SXM4-80GB (2.0\,TB/s) and an RTX A6000 (48\,GB, 0.77\,TB/s), each running SGLang v0.4.2.post3 and v0.5.22.
The main synthetic grid uses CodeLlama-7B-Instruct~\citep{roziere2024codellama}, and \autoref{sec:eval-models} repeats it on five more models; all runs use bf16 and greedy decoding. The real-data benchmarks use Qwen2.5-7B-Instruct~\citep{qwen2025qwen25} with SGLang v0.5.22, on the A100 except for the continuous-batching study (\autoref{tab:cb}), which runs on the A6000. The default number of generated tokens is 128 per question, unless stated otherwise.

\paragraph{Data.}
Following \citet{juravsky2024hydragen}, documents are drawn from \emph{War and Peace}~\citep{warandpeace} and questions are generated per document.
Each document is a different slice of the book, chosen by a random seed, and the $i$-th question comes from a fixed template, ``Summarize the main point of paragraph $i$.'', so all questions of a document have nearly the same length.
The main grid uses 8k- and 16k-token documents with 64, 128 or 256 questions per document.

\paragraph{Real long-document benchmarks.}
We also use two public benchmarks in which many questions share a long document.
\textbf{QuALITY}~\citep{pang2022quality} (v1.0.1, dev split) has 2{,}086 four-way multiple-choice questions about 115 articles, 18 questions per article on average, with articles of 5.7k tokens on average.
Our prompt asks for the letter of the answer, with at most 16 answer tokens, and we report accuracy.
\textbf{LooGLE}~\citep{li2024loogle} (short-dependency QA, test split) has 1{,}761 free-form questions about 96 of its 105 Wikipedia documents that fit a 32k-token context, 18 questions per document on average, with documents of 19.9k tokens on average.
We use its official prompt and answer budget of 300 tokens, and report token recall, the fraction of the reference answer's words that appear in the model's answer, which tracks human judgments of instruction-following models better than F1~\citep{adlakha2024evaluating}. 
In both, the shared document is the instruction plus the document, each question is one stream, and decoding is greedy until the end-of-sequence token or the answer budget.
Calibration uses questions outside the scored set: QuALITY's training split and LooGLE's long-dependency questions.

\paragraph{Methodology.}
Every comparison is \emph{paired}.
A \emph{seed} is the random seed that draws one instance of a workload, that is, its documents and questions; the same seed is timed under every configuration on the same GPU.
We measure \emph{wall time}: the time to answer all questions of one seed, from submitting them until the last answer is returned, with the document already cached.
Each workload uses six evaluation seeds, run in two passes with the configuration order reversed, plus a warm-up seed and a calibration seed.
PP's $P$ is chosen on the calibration seed.
We report the geometric mean of per-seed wall-time ratios with a 95\% bootstrap confidence interval.
Stock SGLang is driven through its offline \texttt{Engine} API in the same process. Before timing, a one-token request per document fills the radix cache; the $Q$ requests (document plus question, already tokenized) are then submitted in a single \texttt{generate} call, with SGLang's default limit on running requests. PP runs on the same \texttt{ModelRunner} through our engine, with the documents prefilled before timing. Each workload uses the same static memory fraction for every configuration (0.8 by default; 0.88 in the model sweep and some A6000 runs). For every configuration, the timed wall time includes question prefill and all decode steps, and excludes tokenization of the inputs.
The real-data benchmarks are timed differently, because their documents are part of the work: each pass starts with an empty prefix cache, and the wall time covers document and question prefill and all decode steps for the whole benchmark. Stock SGLang receives one request per question, all in one \texttt{generate} call, so its radix cache can share a document's prefill among that document's questions. Every configuration first runs a warm-up on the calibration questions, then runs one untimed pass to capture its CUDA graphs. We report the mean of two passes, and all configurations of a table run back to back on one GPU.

\paragraph{Memory.}
We do not compare whole-GPU memory between PP and stock SGLang, because it does not reflect the workload.
SGLang reserves its KV-cache pool when it starts, as a fixed fraction of GPU memory (\texttt{mem\_fraction\_static}).
PP and stock run with the same value in every comparison, so \texttt{nvidia-smi} reports nearly the same memory for both.
Our HuggingFace prototype allocates memory on demand, so for it we report peak GPU memory: PP uses less memory than standard prompting on SQuAD~\citep{rajpurkar2016squad}, QuAC~\citep{choi2018quac}, and DROP~\citep{dua2019drop} (\autoref{tab:mem-table} in \autoref{sec:appendix_material}), and fewer questions per batch lower it further (\autoref{tab:quac-batch-memory} in \autoref{sec:eval-proto}).

\subsection{Speedup over a production engine}
\label{sec:eval-main}

\begin{figure}[t]
\centering
\includegraphics[width=\columnwidth,trim=0 0 0 17,clip]{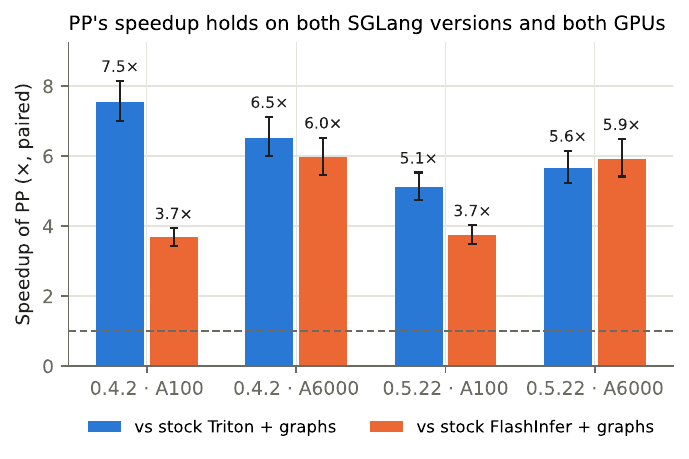}
\caption{Speedup of Parallel Prompting over stock SGLang on 8k/16k-token documents with 64--256 questions. Error bars are 95\% bootstrap CIs over 36 paired runs per setting. PP wins all 144 runs.}
\label{fig:sglang-portability}
\end{figure}

\begin{table}[t]
\centering
\small
\caption{Wall time (s, mean of 6 seeds) on SGLang v0.4.2 with an A100-80GB, and PP's speedup over the faster stock configuration. Wall time is the time to answer all $Q$ questions about one document, which is already cached (\autoref{sec:eval-setup}). FI = FlashInfer; all configurations use CUDA graphs.}
\label{tab:walltime}
\begin{tabular}{@{}rrrrrr@{}}
\toprule
Doc & $Q$ & FI & Triton & PP ($P$) & Speedup \\
\midrule
8k  & 64  & 9.76  & 18.19  & 3.54 (16)  & 2.72$\times$ \\
8k  & 128 & 16.19 & 34.44  & 4.66 (32)  & 3.47$\times$ \\
8k  & 256 & 32.26 & 68.07  & 8.30 (32)  & 3.89$\times$ \\
16k & 64  & 16.17 & 32.61  & 5.29 (16)  & 3.06$\times$ \\
16k & 128 & 30.61 & 63.55  & 6.80 (32)  & 4.50$\times$ \\
16k & 256 & 59.99 & 126.15 & 12.31 (32) & 4.87$\times$ \\
\bottomrule
\end{tabular}
\end{table}

\autoref{fig:sglang-portability} summarizes the main grid.
Against stock FlashInfer, PP is 3.7$\times$ faster on the A100 (v0.4.2 and v0.5.22) and 5.9--6.0$\times$ faster on the A6000.
Against stock Triton, which shares PP's attention backend, it is 5.1--7.5$\times$ faster.
PP wins all 144 paired runs.
The ranges do not overlap: on the v0.4.2 A100 setting, the slowest PP run beats the fastest stock run of the same workload by 2.46--4.76$\times$ (one-sided sign test against the best competitor, $p = 1.5 \times 10^{-11}$).
\autoref{tab:walltime} gives the absolute times.
For example, PP answers 256 questions about a 16k-token document in 12.3\,s, where stock SGLang needs 60.0\,s.

\subsection{Which workloads benefit}
\label{sec:eval-scaling}

\begin{figure*}[t]
\centering
\begin{minipage}[t]{0.66\textwidth}
\includegraphics[width=\linewidth,trim=0 0 0 16,clip]{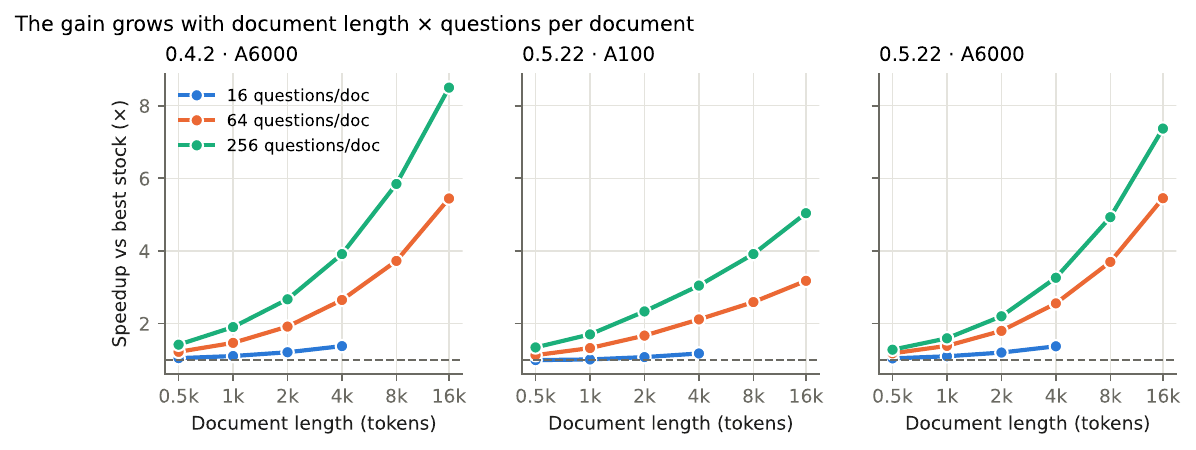}
\end{minipage}\hfill
\begin{minipage}[t]{0.32\textwidth}
\includegraphics[width=\linewidth,trim=0 0 142 18,clip]{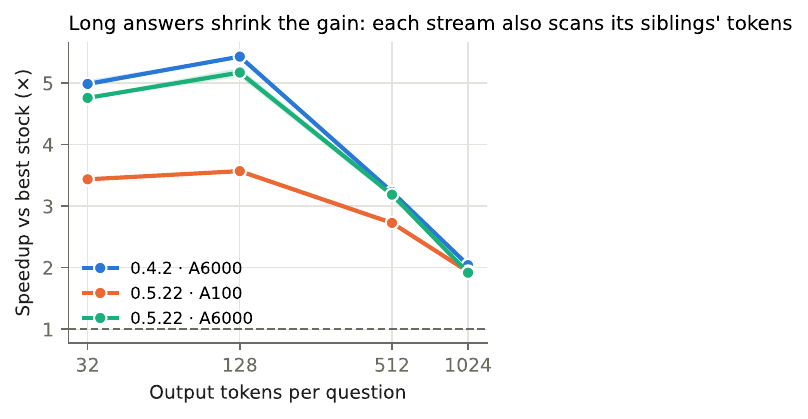}
\end{minipage}
\caption{\textbf{Left:} speedup over the faster stock configuration as document length and questions per document grow (128 output tokens). \textbf{Right:} speedup as answers get longer (8k-token document, 128 questions).}
\label{fig:sglang-scaling}
\end{figure*}

\paragraph{Document length and question count.}
The gain grows with the amount of document KV that stock decoding re-reads (\autoref{fig:sglang-scaling}, left).
With 16 questions about a 512-token document, PP and stock SGLang tie (0.99--1.04$\times$).
With 256 questions about a 16k-token document, PP is 5.0$\times$ faster on the A100 (v0.5.22) and 7.4--8.5$\times$ faster on the A6000.

\paragraph{Output length.}
Long answers shrink the gain (\autoref{fig:sglang-scaling}, right).
From 128 to 1024 output tokens, the speedup falls from 3.6$\times$ to 1.9$\times$ on the A100 (v0.5.22) and from 5.2--5.4$\times$ to 1.9--2.0$\times$ on the A6000, though PP stays faster.
We attribute this to our interleaved layout.
A request's KV row also holds its sibling streams' generated tokens, so at answer length $t$ each stream scans about $P \cdot t$ extra keys that the mask then discards.

\paragraph{Many documents per batch.}
Batching requests from several documents increases the gain.
On v0.5.22 with an A100, 4k-token documents and 64 questions each, PP is 2.1$\times$ faster with one document per batch, 2.9$\times$ with four and 3.3$\times$ with sixteen.
On the A6000, the speedup is 2.6--2.7$\times$ with one document and 3.4--3.7$\times$ with four; sixteen documents do not fit in its KV pool (\autoref{fig:sglang-groups} in \autoref{app:sglang}).

\subsection{Choosing $P$}
\label{sec:eval-p}

\begin{figure*}[t]
\centering
\includegraphics[width=\textwidth,trim=0 0 0 16,clip]{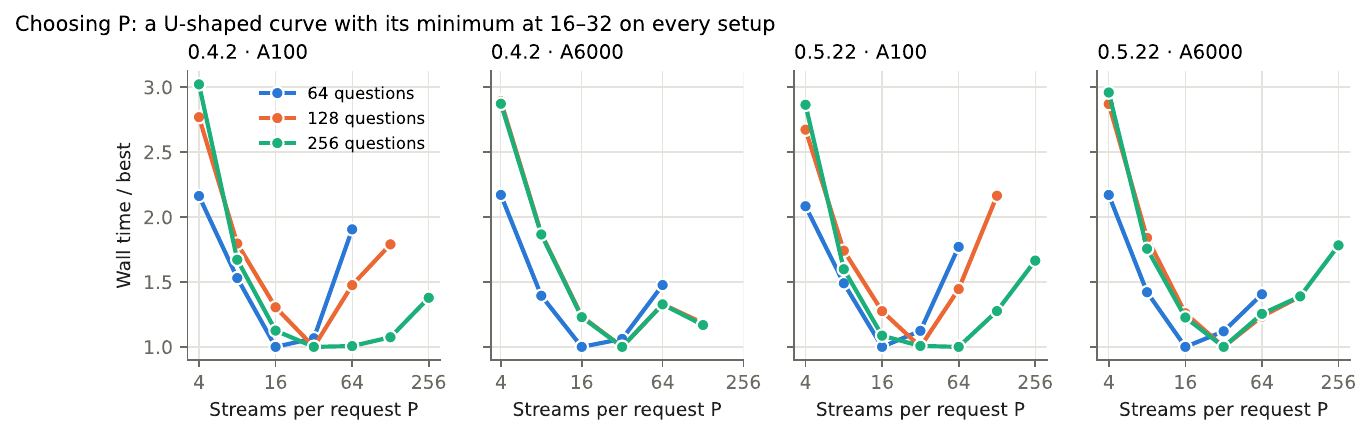}
\caption{Wall time versus streams per request $P$ on the calibration seed, normalized to the best $P$ (16k-token documents). The curve is U-shaped, with its minimum at $P$ = 16--64 on every setting.}
\label{fig:sglang-p}
\end{figure*}

Wall time is U-shaped in $P$, as \autoref{eq:thr} predicts (\autoref{fig:sglang-p}).
Small $P$ leaves the document re-read many times.
Large $P$ makes each stream scan many masked sibling tokens and lengthens each step.
The best $P$ is 16 for 64 questions and 32 for 128 or 256 questions, on both GPUs and both versions; the one exception is $P=64$ for 256 questions on v0.5.22 with the A100.
Near the optimum the curve is flat: with 256 questions about a 16k-token document on the A100, $P = 32$ and $P = 64$ differ by less than 1\%.
A fixed default of $P = 32$ stays within about 12\% of the best $P$ in every setting.

\subsection{Real long-document benchmarks}
\label{sec:eval-real}

\begin{table}[t]
\centering
\small
\caption{QuALITY and LooGLE on an A100-SXM4-80GB with Qwen2.5-7B-Instruct and SGLang v0.5.22, every configuration with the radix prefix cache. Time (s) is the wall time for the whole benchmark from an empty cache, including document prefill (mean of two passes, all rows on one GPU). LooGLE is scored by token recall; token F1 is 17.29, 17.41 and 17.36.}
\label{tab:real}
\setlength{\tabcolsep}{3pt}
\begin{tabular}{@{}lrrrr@{}}
\toprule
 & \multicolumn{2}{c}{QuALITY} & \multicolumn{2}{c}{LooGLE} \\
\cmidrule(lr){2-3}\cmidrule(l){4-5}
Config & Time & Acc & Time & Recall \\
\midrule
Stock FlashInfer & 133.2 & 76.13 & 403.6 & 83.41 \\
Stock Triton & 155.9 & 76.22 & 568.0 & 83.77 \\
PP & 73.6 & 75.89 & 508.3 & 83.61 \\
\bottomrule
\end{tabular}
\end{table}

\autoref{tab:real} runs the configurations of \autoref{sec:eval-setup} on two real benchmarks, timing the whole benchmark from an empty cache.
On QuALITY, whose answers are short, PP answers all 2{,}086 questions in 73.6\,s, 1.81$\times$ faster than stock FlashInfer and 2.12$\times$ faster than stock Triton.
On LooGLE, PP is 1.12$\times$ faster than stock Triton, which uses the same attention kernels, but 1.26$\times$ slower than stock FlashInfer.
LooGLE's answers are long, 234 tokens on average, which shrinks PP's gain because of the interleaved layout (\autoref{sec:eval-scaling}).
And on these 20k-token documents, FlashInfer makes stock SGLang 1.41$\times$ faster than Triton, a kernel advantage that PP, which runs only on Triton, does not get.

\paragraph{Continuous batching.}
Continuous batching can only help when some requests finish earlier than others while more requests wait. That does not happen in either benchmark above: every QuALITY answer runs to its 16-token budget, and each LooGLE batch already holds all of its requests, so none are waiting and PP+CB matches PP there.

\begin{table}[t]
\centering
\small
\caption{Continuous batching on QuALITY in the chat template (Qwen2.5-7B-Instruct, RTX A6000, $P=4$, 512-token budget, mean of two passes). \emph{Uniform}: every answer is the letter (2 tokens). \emph{Mixed}: every other article asks for an explanation first (87 tokens on average, at most 169).}
\label{tab:cb}
\setlength{\tabcolsep}{3pt}
\begin{tabular}{@{}lrrrr@{}}
\toprule
 & \multicolumn{2}{c}{Uniform} & \multicolumn{2}{c}{Mixed} \\
\cmidrule(lr){2-3}\cmidrule(l){4-5}
Config & Time & Steps & Time & Steps \\
\midrule
PP & 161.0 & 752 & 356.5 & 6{,}160 \\
PP+CB & 161.0 & 752 & 292.1 & 4{,}176 \\
\midrule
Speedup & \multicolumn{2}{c}{1.00$\times$} & \multicolumn{2}{c}{1.22$\times$} \\
\bottomrule
\end{tabular}
\end{table}

To isolate the effect, we rerun QuALITY in the model's chat template, so that answers end with the end-of-sequence token, in two versions (\autoref{tab:cb}).
When every answer is the letter alone, all requests finish on the same step, and PP+CB takes exactly as long as PP.
When every other article asks for an explanation before the letter, PP waits in each batch for its longest answer, while PP+CB refills the freed places with waiting requests: it needs 32\% fewer decode steps and is 1.22$\times$ faster, with accuracy within 0.2 points (75.55 vs.\ 75.36).
On the mixed workload, this cuts decode steps from 6{,}160 to 4{,}176.

\paragraph{Answers.}
Accuracy and recall agree within 0.33 points on QuALITY and 0.36 on LooGLE across all configurations.
On SQuAD v1.1, PP's EM and F1 also match both stock backends, and every paired 95\% confidence interval on the F1 difference contains zero (\autoref{sec:eval-quality}).

\subsection{Other model sizes}
\label{sec:eval-models}

\begin{table}[t]
\centering
\footnotesize
\caption{PP speedup across models on SGLang v0.5.22 (paired geometric mean over 6 seeds per workload, 128 output tokens). KV/token is the KV-cache size per token in bf16. Workloads are 8k and 16k-token documents with 64 and 256 questions, except where noted.}
\label{tab:models}
\setlength{\tabcolsep}{2pt}
\begin{tabular}{@{}lcrcccc@{}}
\toprule
Model & Attn. & KV/tok & GPU & vs FI & vs Triton & Wins \\
\midrule
Qwen2.5-1.5B & GQA & 28\,KB & A6000 & 1.28$\times$ & 1.42$\times$ & 12/24 \\
Qwen2.5-7B & GQA & 56\,KB & A6000 & 1.32$\times$ & 1.51$\times$ & 24/24 \\
Qwen2.5-14B & GQA & 192\,KB & A6000 & 1.87$\times$ & 2.08$\times$ & 24/24 \\
Qwen2.5-32B$^\dagger$ & GQA & 256\,KB & A100 & 1.26$\times$ & 1.49$\times$ & 12/12 \\
CodeLlama-7B$^\ddagger$ & MHA & 512\,KB & A6000 & 5.91$\times$ & 5.65$\times$ & 36/36 \\
CodeLlama-13B$^\dagger$ & MHA & 800\,KB & A6000 & 3.54$\times$ & 3.42$\times$ & 12/12 \\
\bottomrule
\end{tabular}
\par\vspace{2pt}
{\footnotesize $^\dagger$8k documents only (KV pool). $^\ddagger$Main grid of \autoref{sec:eval-main}, including 128 questions.}
\end{table}

\autoref{tab:models} repeats the main grid on five more models, including Qwen2.5~\citep{qwen2025qwen25}, which uses grouped-query attention (GQA).
The gain tracks how much KV each stock decode step re-reads.
With full multi-head attention, CodeLlama stores 512--800\,KB of KV per token, and PP is 3.4--5.9$\times$ faster.
GQA shrinks the KV per token, so stock decoding re-reads far less; with 28--56\,KB per token, Qwen2.5-1.5B and 7B gain 1.3--1.5$\times$.
On Qwen2.5-1.5B, the smallest KV in our study, PP is 1.6--1.9$\times$ faster with 256 questions per document  but 5--7\% slower than stock FlashInfer with 64; there, re-reading a 28\,KB-per-token document is cheap, and PP's extra masked keys cost more than they save.
CodeLlama-13B on a 48\,GB GPU leaves a KV pool of only about 20k tokens for stock SGLang, which must then run the batch in small pieces.
Its 3.5$\times$ therefore reflects PP's smaller KV footprint as well as its kernel savings.

The larger Qwen models follow the same trend: Qwen2.5-14B wins all 24 paired runs (1.87$\times$ over FlashInfer, from 1.45$\times$ at 8k/64 to 2.48$\times$ at 16k/256), helped by a tight 76k-token KV pool on 48\,GB, and Qwen2.5-32B on an A100 wins all 12 at 1.26$\times$. For every model, the speedup grows with the number of questions, and for the four models run at both document lengths it also grows with document length. 

\subsection{HuggingFace Implementation}
\label{sec:eval-proto}

Along with the SGLang implementation, we built Parallel Prompting on HuggingFace Transformers~\citep{wolf2020transformers}, with a dense stream mask passed to the model.
The prototype allowed comparisons with methods that have no SGLang implementation, including Hydragen~\citep{juravsky2024hydragen} and batch prompting, and an evaluation of answer quality on reading-comprehension tasks.
With 512-token documents the prototype matched Hydragen's throughput; at 1024 and 2048 tokens, Hydragen's dedicated prefix kernel was faster (\autoref{fig:oom} in \autoref{app:hf}), which motivated moving Parallel Prompting into an engine with fused kernels.
Running on A100-80GB with Llama 3--8B, the prototype is 3.5$\times$ faster than standard prompting on SQuAD and 5.5$\times$ on DROP at equal F1.

\begin{table}[t]
\centering
\small
\caption{Peak GPU memory of the HuggingFace prototype on QuAC (Llama 3--8B, A100-80GB) as the maximum number of questions per batch, across documents, varies. Standard prompting asks one question per prompt, with batches as large as memory allows; its row comes from a separate run. Speedup is over standard prompting.}
\label{tab:quac-batch-memory}
\setlength{\tabcolsep}{4pt}
\begin{tabular}{@{}lrrrrr@{}}
\toprule
Method & Q/batch & Time (s) & Speedup & Mem (GiB) & F1 \\
\midrule
Standard & 32(max) & 1799 & 1.0$\times$ & 53.6 & 33.99 \\
PP & 8 & 872 & 2.1$\times$ & 16.9 & 33.96 \\
PP & 16 & 677 & 2.7$\times$ & 19.8 & 33.98 \\
PP & 32 & 420 & 4.3$\times$ & 24.7 & 34.05 \\
PP & 64 & 352 & 5.1$\times$ & 33.1 & 33.87 \\
PP & 128 & 342 & 5.3$\times$ & 54.1 & 33.92 \\
\bottomrule
\end{tabular}
\end{table}

\paragraph{Memory.}
The HuggingFace prototype allocates memory on demand, so its peak memory shows how PP's footprint scales with the batch.
\autoref{tab:quac-batch-memory} varies the maximum number of questions per batch on QuAC: memory grows with it, while F1 stays within 0.2 points.
With at most 8 questions per batch, PP uses 16.9\,GiB, less than a third of standard prompting's 53.6\,GiB, and is still 2.1$\times$ faster; with 32, it is 4.3$\times$ faster with 46\% of the memory.
On SQuAD and DROP, at the largest batch that fits, PP's peak memory is 13\% and 34\% lower than standard prompting's (\autoref{tab:mem-table} in \autoref{sec:appendix_material}).

\section{Related Work}
\label{sec:related}

\paragraph{Prefix KV reuse.}
PagedAttention~\citep{kwon2023efficient} in vLLM and RadixAttention~\citep{zheng2024sglang} in SGLang share cached prefix KV across requests, and prompt caching~\citep{gim2024prompt} reuses precomputed KV for prompt modules.
These methods remove redundant prefill and memory, but each request still reads the shared KV on every decode step.
Parallel Prompting builds on prefix caching and targets that remaining read.

\paragraph{Shared-prefix attention.}
Hydragen~\citep{juravsky2024hydragen}, RelayAttention~\citep{zhu2024relayattention}, ChunkAttention~\citep{ye2024chunkattention} and FlashInfer's cascade attention~\citep{ye2025flashinfer} split attention into a shared-prefix part, computed for many requests as one matrix--matrix product, and a per-request suffix part, merged afterwards.
Bifurcated attention~\citep{athiwaratkun2024bifurcated} applies the same split to many samples from one context.
DeFT~\citep{yao2025deft} extends this to tree-structured sharing, and FastTree~\citep{pan2025fasttree} groups queries that share a radix-tree prefix inside SGLang and runs them through dedicated Triton kernels.
These approaches need a dedicated kernel and a split or tree-structured KV layout.
Parallel Prompting instead changes how requests are formed: the shared-prefix reuse comes from packing streams into one request, so it reuses the engine's existing extend kernel, KV format and CUDA graphs, with a mask added.
We expect the two ideas to be complementary, since a split-prefix kernel could serve each packed request, but we have not measured the combination.

\paragraph{Packing and batch prompting.}
Prepacking~\citep{zhao2024prepacking} packs prompts of different lengths into one sequence with a block-diagonal mask, for prefill only.
SGLang's multi-item scoring packs several candidates after a shared query, with a mask that keeps the items apart, also for prefill only.\footnote{\url{https://github.com/sgl-project/sglang/pull/10979}}
PDOS~\citep{yu2025pdos} decodes parallel reasoning branches inside one sequence with a tree mask and shared position ids, and concurrent work, IPPD~\citep{glavas2026ippd}, answers several questions over a shared context in one prompt with virtual position ids and an attention mask, implemented in HuggingFace Transformers.
Parallel Prompting applies the same principle to independent questions over a long shared document inside a production engine, with an in-kernel mask that keeps CUDA graphs, and measures the decode-bandwidth savings against the engine's tuned stock configurations.
Batch prompting~\citep{cheng2023batch,lin2024batchprompt} shares the prompt but lets answers interact, which can harm quality.

\paragraph{Faster decoding.}
Speculative decoding~\citep{leviathan2023fast,chen2023accelerating,miao2024specinfer} and multi-token prediction~\citep{cai2024medusa,gloeckle2024better} speed up a single sequence.
They are orthogonal to Parallel Prompting and could run within each stream.
Other work reduces KV traffic through sparsity~\citep{sun2024triforce} or offloading~\citep{sheng2023flexgen}.

\paragraph{Prefix-aware scheduling.}
BatchLLM~\citep{zheng2026batchllm} identifies shared prefixes globally in large offline batches and schedules requests that share them together.
Such schedulers increase prefix sharing; Parallel Prompting reduces the cost of each shared decode step, so the two are complementary.

\section{Limitations and Conclusion}
\label{sec:conclusion}

\paragraph{Limitations.}
Our engine targets batch and offline workloads, in which the requests that share a document are known in advance; serving requests as they arrive requires integration with an online scheduler, which we leave to future work.
The implementation also supports only a subset of the serving engine's features, such as one attention backend, a single GPU and greedy decoding.
The benefit depends on the workload: it is largest when many questions share a long document and answers are short, and it shrinks when answers are long or vary widely in length, when the model stores little KV cache per token, or when the baseline uses faster attention kernels. For such workloads, an engine can fall back to standard batching.
Finally, Parallel Prompting changes the order of floating-point operations, so its outputs are not bitwise identical to standard decoding, although task accuracy matches it within a fraction of a point in our experiments.

\paragraph{Conclusion.}
Prefix caching stops recomputing a shared document but still re-reads it once per question at every decode step.
Parallel Prompting decodes many questions as independent streams of one request, so each step reads the document once per request instead of once per question.
Parallel prompting is 3.7--6.0$\times$ faster than stock SGLang with FlashInfer on long-document, many-question workloads with CodeLlama-7B, and 1.3--1.9$\times$ faster on grouped-query-attention Qwen2.5 models from 7B to 32B.

\section*{Acknowledgements}
This work used resources available through the National Research Platform (NRP) at the University of California, San Diego. NRP has been developed, and is supported in part, by funding from National Science Foundation, from awards 1730158, 1540112, 1541349, 1826967, 2112167, 2100237, and 2120019, as well as additional funding from community partners.

\FloatBarrier
\bibliography{references}
\bibliographystyle{mlsys2025}

\appendix
\section{Additional SGLang Results}
\label{app:sglang}

\begin{figure}[ht]
\centering
\includegraphics[width=0.8\columnwidth,trim=0 0 0 19,clip]{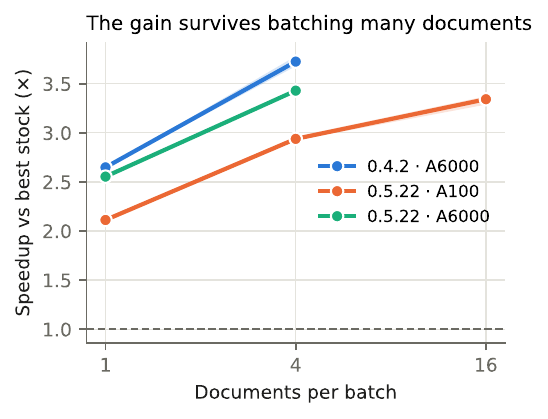}
\caption{Speedup of PP over the faster stock configuration as more documents share a batch (4k-token documents, 64 questions each, 128 output tokens). Sixteen documents do not fit in the A6000's KV pool.}
\label{fig:sglang-groups}
\end{figure}

\autoref{fig:sglang-groups} plots the multi-document results of \autoref{sec:eval-scaling}.

\section{Answer Quality}
\label{sec:eval-quality}

With exact arithmetic, \autoref{eq:streammask} and per-stream positions give each stream exactly the attention inputs of a single-question request.
In bf16, however, the different packing changes the order of floating-point reductions.

\begin{table}[t]
\centering
\small
\caption{SQuAD v1.1 dev (10{,}570 questions, 48 articles split into 71 chunks of at most 8k tokens) on SGLang v0.5.22 with CodeLlama-7B-Instruct on an RTX A6000. Answer time excludes document prefill for all configurations. $\Delta$F1 is PP minus stock, with a paired bootstrap 95\% CI over questions.}
\label{tab:squad}
\setlength{\tabcolsep}{4pt}
\begin{tabular}{@{}lrrcr@{}}
\toprule
Config & EM & F1 & $\Delta$F1 [95\% CI] & Time (s) \\
\midrule
PP ($P{=}32$) & 21.51 & 45.14 & -- & 237.6 \\
FlashInfer & 21.50 & 45.25 & $-0.11$ [$-.25$, $.03$] & 406.5 \\
Triton & 21.55 & 45.18 & $-0.04$ [$-.20$, $.12$] & 409.2 \\
\bottomrule
\end{tabular}
\end{table}

To measure task quality, we run SQuAD v1.1~\citep{rajpurkar2016squad} through the SGLang implementation.
Each Wikipedia article is the shared document, split into chunks of at most 8k tokens (5.5k on average), and its questions become the streams: 149 per chunk on average and up to 766.
All configurations use the same instruction prompt, greedy decoding and at most 32 answer tokens.
\autoref{tab:squad} shows that PP's EM and F1 match both stock backends: every paired confidence interval on the difference contains zero.
On this realistic workload, with documents shorter than in \autoref{sec:eval-main}, PP generates the answers 1.71$\times$ faster.
Absolute EM is low for every configuration because the model answers with a short sentence that contains the span rather than the bare span.

On our HuggingFace prototype with Llama 3--8B, PP matches standard prompting's F1 on SQuAD~\citep{rajpurkar2016squad}, QuAC~\citep{choi2018quac} and DROP~\citep{dua2019drop}, while batch prompting loses accuracy (\autoref{tab:empirical-table-horizontal} in \autoref{app:hf}); on QuAC, the prototype scores only the 81\% of questions whose dialogue history fits in memory.

\section{Theoretical Analysis}
\label{app:theory}

This appendix analyzes the efficiency gains of Parallel Prompting in LLM inference. We first recall Amdahl's Law for parallel algorithms and then derive the speedup and throughput improvements of our approach.

Amdahl's Law~\citep{amdahl1967validity} gives the speedup of a task when a portion of it is parallelized:
\begin{equation}
S(n) = \frac{1}{(1 - p) + \frac{p}{n}}
    \label{eq:Amdahl}
\end{equation}
where  $S(n)$ is the speedup with $n$   processors,
 $p$  is the fraction of the task that can be parallelized,
 $1-p$ is the fraction that remains serial.
The overall speedup is therefore limited by the serial portion of the task. As $n$ increases, the speedup approaches $\frac{1}{1-p}$, so returns diminish unless $p$ is close to 1.

In LLM inference, traditional methods process each query sequentially, which is inefficient because decoding processing is serial.Parallel Prompting decodes multiple queries in one sequence, so attention over the shared prefix becomes a matrix-matrix product. We measure throughput as the number of queries (prompts) processed, each to a full output completion, per unit time.

\begin{proposition}
    \label{thm:theorem3}
Consider inference on $N$ independent queries using (a) standard batch processing and (b) parallel prompting (packing all queries as independent subsequences in a single sequence with attention masking).

Let $T_\text{batch} = T_\text{setup} + N \cdot T_{\mathrm{MV}}$ be the wall-time for a batch (with matrix-vector attention), and $T_\text{parallel} = T_\text{setup} + T_{\mathrm{MM}}$ for parallel prompting (with matrix-matrix attention). Then, the respective throughput values are:
\begin{align}
\text{Throughput}_\text{batch} = \frac{N}{T_\text{batch}}, \\
\text{Throughput}_\text{parallel} = \frac{N}{T_\text{parallel}}
    \label{eq:statement1}
\end{align}
and
\begin{align}
\frac{
    \text{Throughput}_\text{parallel}
}{
    \text{Throughput}_\text{batch}
}
=
\frac{
    T_\text{batch}
}{
    T_\text{parallel}
}
=
\frac{
    T_\text{setup} + N T_{\mathrm{MV}}
}{
    T_\text{setup} + T_{\mathrm{MM}}
}
    \label{eq:statement11}
\end{align}
where $T_{\mathrm{MV}}$ is per-query wall-time for the matrix-vector attentions, and $T_{\mathrm{MM}}$ is wall-time for the matrix-matrix product in the attention.

Since $T_{\mathrm{MM}} \geq T_{\mathrm{MV}}$, the ratio is at most $N$, and it approaches this upper bound only if $T_\text{setup}$ is negligible and $T_{\mathrm{MM}}$ stays close to $T_{\mathrm{MV}}$.
\end{proposition}

\textbf{Assumptions.}
\begin{itemize}
    \item The model and hardware support this masking and packing; $T_{\mathrm{MV}}$ and $T_{\mathrm{MM}}$ are measured compatibly.
    \item Setup time is equal for standard batch processing and parallel prompting.
    \item $N$ is small enough to avoid exceeding hardware or memory limits for both methods.
\end{itemize}

In practice $T_{\mathrm{MM}}$ grows with $N$: each stream also scans its sibling streams' masked keys (\autoref{sec:eval-scaling}). The measured gains, at most 8.5$\times$ (\autoref{sec:eval}), are therefore far below $N$, and the bound is not attained.

\subsection{Throughput Maximization by Balancing Attention Parallelism and Batch Size}

Batching is an important technique for raising throughput in LLM inference. In batched decoding, each forward pass processes the latest token of many sequences instead of one. This raises the arithmetic intensity of transformer components such as the multilayer perceptron (MLP) blocks and lets them use hardware-friendly matrix multiplications.

The arithmetic intensity of attention does not benefit from batching, because each sequence has its own key and value matrices. While the other components use tensor cores during batched decoding, attention must still be computed as many independent matrix-vector products. Parallel Prompting aims to raise the arithmetic intensity of attention.

\begin{proposition}[Throughput Maximization]
\label{prop:throughput-max}
Let $P$ be the number of streams per request (number of independent queries packed into a sequence for matrix-matrix attention), $B$ the batch size (number of such sequences processed in parallel), and $P\cdot B \leq S^*$ a hardware resource constraint (e.g., total token capacity).

Let $T_{\mathrm{attn}}(P)$ denote the attention computation cost (function of $P$), and $T_{\mathrm{mlp}}(PB)$ denote the MLP/other cost (a function of the $PB$ tokens per step).

Then, the throughput (queries per unit time) satisfies:
\begin{align}
\mathrm{Throughput}(P, B) = \frac{P \cdot B}{T_{\mathrm{attn}}(P) + T_{\mathrm{mlp}}(PB)}
    \label{eq:proposition}
\end{align}
and maximal throughput is achieved at
\begin{align}
(P^*, B^*) = \operatorname*{arg\,max}_{P \cdot B \leq S^*} \frac{P \cdot B}{T_{\mathrm{attn}}(P) + T_{\mathrm{mlp}}(PB)}
    \label{eq:proposition2}
\end{align}
where the per-stream attention cost $T_{\mathrm{attn}}(P)/P$ generally falls with $P$ up to a hardware limit (then rises), and $T_{\mathrm{mlp}}(PB)$ grows sublinearly in $PB$ up to a limit.

\end{proposition}

\textbf{Assumptions.}
\begin{itemize}
    \item $P$ queries are packed per prompt and $B$ prompts per batch, with $PB \leq S^*$ (a resource or hardware constraint).
    \item $T_\mathrm{attn}(P)$ and $T_\mathrm{mlp}(PB)$ are the attention and MLP wall times per step.
    \item The per-token costs $T_\mathrm{attn}(P)/P$ and $T_\mathrm{mlp}(PB)/(PB)$ are nonincreasing up to hardware limits, and nonmonotone beyond them.
\end{itemize}

The maximizing pair $(P^*, B^*)$ is found by balancing optimal matrix-matrix utilization for attention and optimal batch size for MLP efficiency. We expect the throughput function to be quasi-concave in $(P, B)$ under natural hardware scaling assumptions for transformer kernels; we do not prove this, but the U-shaped wall-time curves in \autoref{fig:sglang-p} are consistent with it. The theoretical maximum could exist at an interior point determined by hardware and model specifics, and is not achieved by maximizing either $P$ or $B$ alone.

\section{Further Results of the HuggingFace Prototype}
\label{app:hf}
This appendix gives further results from our first implementation of Parallel Prompting, built on HuggingFace Transformers~\citep{wolf2020transformers} (\autoref{sec:eval-proto}).

We evaluate the prototype in two sets of experiments:
(1) controlled scaling studies on small and medium-sized models using synthetic data, and
(2) a downstream task evaluation on reading comprehension datasets using Llama 3–8B.
All experiments run on a single NVIDIA A100-80GB GPU with PyTorch implementations built on HuggingFace Transformers~\citep{wolf2020transformers}.

\subsection{Scaling Experiments}

\paragraph{Setup.}
Following~\citet{juravsky2024hydragen}, we construct synthetic datasets with varying document lengths, numbers of unique documents, and numbers of queries.
Document content is drawn from a subset of \emph{War and Peace}~\citep{warandpeace}.
We perform all scaling studies on CodeLlama-7B-Instruct~\citep{roziere2024codellama} and Sheared-LLaMA-1.3B~\citep{xia2024shearedllamaacceleratinglanguage} to enable controlled analysis under constrained compute.

\paragraph{Memory Constraints and Throughput Under Increasing Context Length.}
We first examine memory usage and throughput as the number of queries and the shared-context length increase.
\autoref{fig:oom} summarizes the results.
Throughput falls with longer documents for both methods. The prototype matches Hydragen at 512 tokens, but Hydragen's dedicated prefix kernel is faster at 1024 and 2048 tokens, which motivates moving Parallel Prompting into an engine with fused attention kernels (\autoref{sec:impl}).

\begin{table}[t]
\centering
\footnotesize
\caption{Synthetic throughput (tokens/s) of the prototype and Hydragen, 8 documents, CodeLlama-7B-Instruct, A100-80GB.}
\label{fig:oom}
\begin{tabular}{@{}llccc@{}}
\toprule
Questions & Method & 512 & 1024 & 2048 \\
\midrule
$8\times128$ & Hydragen & 2479 & 2161 & 1743 \\
$8\times128$ & Ours & 2528 & 1995 & 1355 \\
\bottomrule
\end{tabular}
\end{table}

\subsection{Case Study: Question Answering Performance}

We evaluate our method on downstream reading comprehension tasks to measure its end-to-end effect on answer quality and generation speed.
We use Llama 3–8B~\citep{grattafiori2024llama3herdmodels} and measure F1 scores (standard for QA) on SQuAD~\citep{rajpurkar2016squad}, QuAC~\citep{choi2018quac}, and DROP~\citep{dua2019drop}.

As shown in \autoref{tab:empirical-table-horizontal}, Parallel Prompting keeps the F1 of standard prompting and is 3.5$\times$, 8.1$\times$ and 5.5$\times$ faster on SQuAD, QuAC and DROP, while sequential batch prompting loses F1. On QuAC, the Parallel time covers only the 5{,}951 of 7{,}354 questions whose dialogue history fits in memory, so the QuAC ratio overstates the like-for-like speedup.

\section{Additional Prototype Results}

\label{sec:appendix_material}

We use smaller models and synthetic data for the analytical and ablation studies than for the downstream evaluations. Large models such as Llama 3-8B are computationally expensive, which makes extensive ablation and scaling experiments over many parameters hard to run. Smaller models and synthetic datasets let us vary batch size, prefix length and number of queries systematically and isolate the effect of our method under controlled conditions. The synthetic studies show how throughput and memory change with batch size, prefix length and number of queries, and the downstream benchmarks measure end-to-end quality and speed.

\begin{table*}[t]
\caption{Comparison of generation time and performance for downstream tasks with different methods with Llama 3 8B model on A100-80G. Standard, SeqBatch and Parallel are rerun with the original scripts; Hydragen and vLLM[0.6.4] are from the original runs. \autoref{sec:eval-main} compares against current SGLang. Std is the across-run standard deviation of the time where available. $^*$QuAC Parallel and SeqBatch score only the questions that fit in memory (5{,}951 and 2{,}899 of 7{,}354). F1 is computed as the harmonic mean of precision and recall in extractive QA.}
\vspace{6pt}
\label{tab:empirical-table-horizontal}
\centering
\begin{tabular}{l|ccc|ccc|ccc}
\hline
\multirow{2}{2em}{\textbf{Method}} & \multicolumn{3}{c|}{\textbf{SQuAD}} & \multicolumn{3}{c|}{\textbf{QuAC}} & \multicolumn{3}{c}{\textbf{DROP}} \\
&\textbf{Times(s)} & \textbf{Std} & \textbf{F1(\%)} & \textbf{Time(s)} & \textbf{Std} & \textbf{F1(\%)} & \textbf{Time(s)} & \textbf{Std} & \textbf{F1(\%)}\\
\hline
Standard  & 597 & -- & 87.2 & 1937 & -- & 34.0 & 647 & -- & 58.1  \\
SeqBatch  &  365 & -- & 84.3 &  346$^*$ & -- & 32.4$^*$ & 843 & -- & 42.6 \\
Hydragen  & 1651 & 20.9 & 87.1 & 1230 & 6.74 & 34.0 &  471 & 3.85 & 58.2 \\
vLLM      &  373 & 0.46 & 87.4 &  897 & 0.57 & 32.8 &  417 & 0.44 & 58.5 \\
\textbf{Parallel}  &  169 & -- & 87.2 &  238$^*$ & -- & 34.2$^*$ & 117 & -- & 58.2 \\
\hline
\end{tabular}
\end{table*}
\paragraph{Memory Usage} \autoref{tab:mem-table} shows that the Parallel method uses less GPU memory than standard prompting on all three datasets. Its memory use depends on the maximum number of questions per batch, which we set as large as memory allows. Reducing it on QuAC lowers memory further while keeping a large speedup over the baseline. \autoref{tab:quac-batch-memory} lists memory, time and F1 for several values.
\begin{table*}[t]
  \caption{Comparison of memory usage with different methods with Llama 3 8B model on A100-80G. These runs are a separate run set from \autoref{tab:empirical-table-horizontal}, so their times differ. Memory is in GiB; the SQuAD values were converted from 55{,}700 and 48{,}600\,MiB.}
   \vspace{6pt}
  \label{tab:mem-table}
  \centering
  \begin{tabular}{ll|p{3em}|p{5em}}
    \hline
    \textbf{Dataset} & \textbf{Method}  & \textbf{Time(s) }& \textbf{Memory(GiB)}\\
    \hline
    \multirow{2}{2em}{\small SQuAD}
    & Standard  & 590    &  54.4  \\
    & Parallel  & 168     & 47.5    \\
        \hline
    \multirow{2}{2em}{\small QuAC}
    & Standard  & 1799     & 53.6   \\
    & Parallel  & 352    & 33.1    \\
        \hline
    \multirow{2}{2em}{\small DROP }
    & Standard  & 654    & 54.3  \\
    & Parallel  & 111   & 36.1   \\
        \hline
  \end{tabular}
\end{table*}

\paragraph{Effect of the Number of Questions.}
We sweep over the number of queries for fixed document and query lengths.
\autoref{tab:parallel sizes1b} shows that throughput improves as the number of parallel queries increases, more for the 1.3B model than for the 7B model.
At small batch sizes, non-attention operations dominate; at large query counts we expect attention over long prefixes to become the bottleneck, and parallel decoding to give larger gains there.
\begin{table*}[t]
\caption{Throughput (tokens/sec) under different batch sizes for parallel generation with Sheared-LLaMA-1.3B and CodeLlama-7B when
$doc\_{len}=512$, $q\_{len}=12$, and $ans\_{len}=5$. }
\vspace{6pt}
\label{tab:parallel sizes1b}
\centering
\begin{tabular}{lccc}
\toprule
$\textbf{Num\_Questions}$ & $\textbf{Batch Size}$ & $\textbf{Throughput-}$$\textbf{1.3B}$ & $\textbf{Throughput-7B}$ \\
\midrule
\multirow{4}{*}{128}
& 1 & 4283 & 1931 \\
& 2 & 4625 & 1843 \\
& 4 & 3654 & 1468 \\
& 8 & 2850 & 1018 \\
\midrule
\multirow{4}{*}{256}
& 1 & 5911 & 2115 \\
& 2 & 6384 & 2250 \\
& 4 & 5748 & 2071 \\
& 8 & 4959 & 1615 \\
\midrule
\multirow{4}{*}{512}
& 1 & 5419 & 1850 \\
& 2 & 6845 & 2214 \\
& 4 & 7725 & 2382 \\
& 8 & 7181 & 2146 \\
\bottomrule
\end{tabular}
\end{table*}

\paragraph{Sequence Length vs. Computation Gains Trade-off.} Our measurements show that throughput increases with batch size or streams per request up to a point, after which the overhead of the longer packed input sequences outweighs the advantage of matrix-matrix attention, as \autoref{prop:throughput-max} predicts. In our SGLang measurements, $P$ = 16--32 is optimal, and $P$ = 64 in one setting (\autoref{sec:eval-p}).

\paragraph{Compatibility with Speculative Decoding.} Parallel Prompting decodes multiple suffixes in lock-step to generate answers to several queries at once, while speculative decoding verifies tokens of a single sequence. The two are distinct but may be complementary: speculative decoding could run within each Parallel Prompting stream, or be adapted to verify multiple shared-prefix continuations in parallel.

\paragraph{Developer Overhead and Practical Adoption.} In many production stacks, the shared-prefix boundary is already explicit: for example, retrieval-augmented generation (RAG) pipelines concatenate retrieved context (prefix) with a question (suffix), and batched APIs group queries under a common header or instruction. In these settings, Parallel Prompting needs only two inputs: (1) the token span (or delimiter) of the shared prefix and (2) a list of per-query suffixes. Our current engine supports only offline workloads (\autoref{sec:conclusion}); online serving needs scheduler support.

\paragraph{Effect of Model Size} We also vary document length for CodeLlama-7b-Inst~\citep{roziere2024codellama} and Sheared-LLaMA-1.3B~\citep{xia2024shearedllamaacceleratinglanguage} (\autoref{tab: doc length}). We therefore also run several configurations with CodeLlama-7b-Inst~\citep{roziere2024codellama} and Sheared-LLaMA-1.3B~\citep{xia2024shearedllamaacceleratinglanguage}.  See~\autoref{tab: doc length} for results.

\begin{table*}[t]
\caption{Comparing the throughput using parallel Batching with CodeLlama-7B and Sheared-LLaMA-1.3B with different lengths of doc length when $q\_{len}=12\|q\_{num}=128\|ans\_{len}=5$ and the number of unique doc content equals 8. Throughput falls by about 30\% for both models as the document length grows from 256 to 1024 tokens.}
\vspace{6pt}
\label{tab: doc length}
\centering
\begin{tabular}{lcccc}
\hline
$doc\_{len}$    & $Throughput($$1.3B$$)(tokens/second)$ & $Throughput(7B)(tokens/second)$\\
\hline
256 & 9514  & 2751\\
512 & 8199 & 2430 \\
1024 & 6592  &1924 \\\hline
\end{tabular}

\end{table*}

\end{document}